\let\oldvec\vec
\documentclass[]{llncs}
\let\vec\oldvec

\usepackage{graphicx}
\usepackage{float}
\usepackage{subfigure}
\def\unotecs#1{\csname unote:#1\endcsname}
\def\unote#1{\leavevmode\hbox to0pt{\vtop{\null \kern-2.5ex\kern\unotecs{#1A}%
   \hbox{$\underbrace{\hskip\unotecs{#1B}}_{\hbox{\unotecs{#1C}}}$}}\hss}}
\def\unotedef#1#2#3#4{%
   \expandafter\def\csname unote:#1A\endcsname{#4}%
   \setbox0=\hbox{#2}\expandafter\edef\csname unote:#1B\endcsname{\the\wd0}%
   \expandafter\def\csname unote:#1C\endcsname{#3}%
}

\graphicspath{{figures/}}

\begin{document}
\title{Multitask Models for Supervised Protest Detection in Texts}
\titlerunning{Protest Detection in Text}
%
\author{Benjamin J. Radford\inst{1}\orcidID{0000-0002-8440-0655}}
\authorrunning{B. Radford}
%
\institute{University of North Carolina at Charlotte, Charlotte NC 28223, USA \\
\email{benjamin.radford@uncc.edu}}
\maketitle              
\begin{abstract}
The CLEF 2019 ProtestNews Lab tasks participants to identify text relating to political protests within larger corpora of news data. Three tasks include article classification, sentence detection, and event extraction. I apply multitask neural networks capable of producing predictions for two and three of these tasks simultaneously. The multitask framework allows the model to learn relevant features from the training data of all three tasks. This paper demonstrates performance near or above the reported state-of-the-art for automated political event coding though noted differences in research design make direct comparisons difficult.

\end{abstract}

{\let\thefootnote\relax\footnotetext{Copyright \textcopyright\ 2019 for this paper by its authors. Use permitted under Creative Commons License Attribution 4.0 International (CC BY 4.0). CLEF 2019, 9-12 September 2019, Lugano, Switzerland.}}

\section{Introduction}

H{\"u}rriyeto{\u{g}}lu et al. propose a competitive lab in which participants are tasked with producing models to automatically identify indicators of protest in cross-country (but monolingual) text corpora~\cite{hurriyetoglu:etal:2019}. With respect to application area, this challenge builds on work done in political science on deriving structured data about political events of interest from unstructured texts (i.e. news). Methodologically, the lab is structured as a competition in which select data are provided to competitors for model training and other data are withheld for evaluation purposes. The tasks themselves fall into the categories of text classification at the document (task 1) and sentence (task 2) levels and semantic role labeling (task 3). 

This paper proceeds by first introducing the three challenge tasks and the provided data. Then the two models are described: one model for tasks 1 and 2 and a second model for tasks 1, 2, and 3. Results for each task are discussed and compared to the published state-of-the-art on similar tasks. Finally, directions for future research are highlighted.

\subsection{Data and Task Description}

The competition comprises three tasks. All three tasks evaluate participants' ability to identify indicators of protest events in English text data. However, the three tasks differ in resolution (document-, sentence-, and word-level data) and in provided data sets. Each task comprises four data sets: \emph{train}, \emph{dev}, \emph{test}, and \emph{China}. As its name implies, the \emph{train} data set is used to train models. The \emph{dev} data set is a validation set provided to participants for model fine-tuning. The \emph{test} data set is out-of-sample and therefore the labels associated with these data are withheld from participants. Similarly, the \emph{China} data set is an out-of-sample set used to evaluate cross-country performance of models. \emph{Train}, \emph{dev}, and \emph{test} contain text data from English-language news stories collected from Indian sources. The \emph{China} data set is composed of English-language news collected from Chinese sources.   Lab participants are able to observe $X$ and $y$, the texts and associated labels, for \emph{train} and \emph{dev}. Participants can only observe $X$, the texts, for \emph{test} and \emph{dev}. 

The role of the \emph{China} set is to evaluate the performance of models in a cross-country setting; more specifically, the \emph{China} data appear similar in form to the \emph{test} data (in that they are English news wire text) but are generated by different underlying data generating processes (DGP). The DGP of the \emph{train}, \emph{dev}, and \emph{test} sets represent Indian political and social processes as well as reporting norms, standards, and laws. The \emph{China} data set represents the same for China.

A small amount of data preprocessing is performed prior to modeling. All non-alphanumeric characters are removed and all whitespace characters (e.g. tabs, newlines, spaces) are replaced with a single space. For tasks 1 and 2, characters are all converted to lowercase.\footnote{For task 3, characters are not converted to lowercase because it would have negatively impacted the performance of the named entity recognition preprocessing step, described later.} All sequences are zero-padded such that every sequence within a given task's corpus is of equal length. The sequence length for each corpus is equivalent to the maximum sequence length observed in that corpus prior to padding (given in the following subsections). This is done to satisfy a software requirement that input sequences are of the same length during model training.

\subsubsection{Document Classification}

Task 1 challenges participants to classify documents, in this case news articles, as one of either relating to a protest or not relating to a protest. Documents in the \emph{train} data set vary in size from 44 words to 1599 words. The mean document length is 312 words. Total data set sizes are given in Table~\ref{tab:datasize}. 

\begin{table}\centering
\caption{Data set size by task.}\label{tab:datasize}
\begin{tabular}{@{\extracolsep{4pt}}lrrr@{}}
\hline\hline
& \multicolumn{1}{c}{Task 1} & \multicolumn{1}{c}{Task 2} & \multicolumn{1}{c}{Task 3} \\
Data set & \multicolumn{1}{c}{(documents)} & \multicolumn{1}{c}{(sentences)} & \multicolumn{1}{c}{(words\footnotemark{})} \\
\hline
Train & 3,429 & 5,884 & 21,873 \\
Dev & 456 & 662 & 3,224 \\
Test & 686 & 1,106 & 6,586 \\
China & 1,800 & 1,234 & 4,387 \\
\hline\hline
\end{tabular}
\end{table}
\footnotetext{Including control words that indicate the beginning and end of sequences.}

\subsubsection{Sentence Classification}

Task 2 is similar to task 1 performed not on the document level but at the sentence level. Given a sentence, the model is tasked to predict whether the sentence describes a protest event or not. Task 2 \emph{train} data set sentences range in length from one word to 150 words and have a mean of length of 24 words.\footnote{In fact, two entries appear to have no words -- they are empty strings. It is unclear if this is a problem with the original data, the download process, or the pre-processing steps.}

\subsubsection{Semantic Role Labeling}

Task 3 differs from tasks 1 and 2 in that it is effectively a multiclass classification problem. Given sentences tokenized at the word level, participants are tasked with identifying sets of words (or phrases) that represent particular roles in the context of a protest. These roles include \emph{triggers}, \emph{locations}, \emph{facilities}, \emph{organizers}, \emph{participants}, \emph{event times}, and \emph{targets}. Tokens are labeled using IOB, \emph{inside, outside, beginning}, tags~\cite{ramshaw:marcus:1995}. Tokens labeled ``O'' are outside of a role tag. Tokens labeled ``B'' represent the beginning of a phrase that is associated with one of the roles. Tokens labeled ``I'' are inside of an identified phrase associated with a role. An example from the \emph{train} dataset is given in Figure~\ref{fig:annotationexample}.

\begin{figure}\centering
\begin{tabular}{@{\extracolsep{8pt}}lllllll@{}}
The & protesters & blocked & the & Rayakottai & road & . \\
O & B-participant & B-trigger & O & B-fname & I-fname & O \\
\end{tabular}
\caption{An example of a role-labeled sentence from the \emph{train} data set. The top row is the provided sentence and the bottom row consists of token-level role annotations.}\label{fig:annotationexample}
\end{figure}

\subsection{Prior Work}

A robust research effort within political science has seen many iterations on techniques for both manual and automatic coding of event records from unstructured texts. Most previous work on automated event coding relies on large dictionaries of terms and phrases organized into known ontological categories. These dictionaries are provided alongside text data to event-coding software that performs pattern matching to identify instances of dictionary phrases within the texts. If those phrases found in the texts match a set of heuristics, the software produces an event record. Protests, the event category of interest here, represent just one ontological category within the CAMEO ontology, the most common of event-coding ontologies in use today~\cite{gerner:etal:2002}. Dictionary-based event coding software includes TABARI~\cite{schrodt:2009} and PETRARCH \cite{schrodt:etal:2014} which have been used in the production of many event data sets including ICEWS~\cite{obrien:2010}, GDELT~\cite{leetaru:schrodt:2013}, and Phoenix~\cite{beieler:2016b}.

While most dictionaries for coding event data are hand-coded by researchers, recent efforts have sought to largely automate the dictionary generation process as well -- a step towards a fully-automated event data pipeline.\footnote{It is arguable that a fully-automated event data pipeline is not desirable -- if one intends to produce structured event records, they likely desire those records to conform to some mental model. Withholding the desired mental model from the event data collection process risks producing records that do not conform to the desired categories or ontology. Therefore, it is difficult to imagine scenarios where fully-unsupervised event data collection is preferable to supervised or semi-supervised event data collection.}  One such effort makes use of distributed word vectors to populate dictionaries given a small input set of exemplar (``seed'') phrases~\cite{radford:2016,radford:2019}. Similar work leverages label propagation to expand a given set of terms and phrases for event coding~\cite{makarov:2018}.

Most recently, supervised learning has been applied to the problem of event identification within text with the goal of producing an end-to-end solution. A neural network technique similar to that presented here was used by Beieler to label sentences according to the Schrodt's \emph{QuadClass} ontology~\cite{beieler:2016,schrodt:2006}. That research assumed the existence of an event in the provided text and tasked a model to classify the event as one of four types; this differs from the task at hand -- to predict event existence versus non-existence and to identify the key actors and actions relevant to a protest event.

\section{Models}

While the model presented here for tasks 1 and 2 differs from the model for task 3, they have several properties in common. Both models are examples of recurrent neural networks (RNNs). RNNs expect time-ordered inputs and are able to model time-dependent sequences by persisting information across time steps. These models differ from traditional autoregressive statistical models in that the lag structure is variable, not pre-determined. Both models are also multitask; that is, each model is trained on examples from two or more of the tasks simultaneously. Finally, the inputs to both models are, at least in part, sequences of words (or tokens). However, prior to the modeling stage, every word has been replaced by a its corresponding word vector. The word vectors are pre-trained on the English Wikipedia corpus using FastText, a neural network language model that leverages both contextual information and sub-word information to produce word vectors~\cite{bojanowski:etal:2016}. Word vectors are real-valued numerical vectors that encode semantic and syntactic relationships between words. The word vector representations of synonyms should be close to one another (where ``closeness'' often means having a high cosine similarity). FastText models words as the combination of sub-word n-grams (letter sequences). The use of pre-trained word vectors has become common for applications in which training a novel word embedding model may be infeasible due to, for example, corpus size or compute resources~\cite{mikolov:etal:2018}. FastText-based vectors were chosen because, unlike word2vec-based vectors, out-of-sample inference can be performed with FastText. If words exist in our corpora that do not exist in the vocabulary that the FastText model was trained on, new word vectors for those out-of-sample words can be derived from the sub-word (i.e. character $n$-gram) information of those out-of-sample words. 

\begin{figure}\centering
\includegraphics[width=\linewidth]{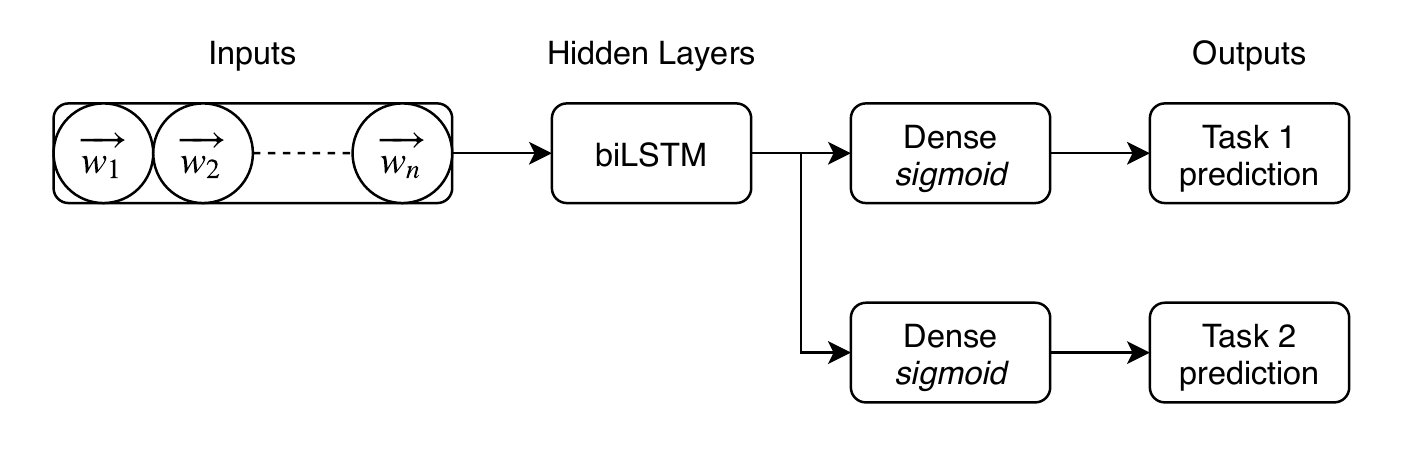}
\caption{Model architecture for tasks 1 and 2. Sequence input is shown on the left for word vectors $\vec{w_1}$ through $\vec{w_n}$.}\label{fig:task12}
\end{figure}

\subsection{Tasks 1 and 2 Model}

The model for tasks 1 and 2 is simple. It takes as input a time-ordered sequence of tokens (i.e. words) of arbitrary length (and possibly padded with zero vectors) and outputs a document-level and sentence-level prediction that the given sequence describes a protest event. The input tokens are length 300 real-valued distributed word vectors derived from the pre-trained FastText model. The models first layer consists of 10 bidirectional long short-term memory (LSTM) RNN cells with no activation function~\cite{hochreiter:schmidhuber:1997}. The layer outputs only the activation values of the cells at the final sequence token -- a $10 \times 1$ real-valued vector. This output connects to two dense, fully-connected layers of size $10 \times 1$ that compute the weighted sum of the 10 activation values from the LSTM's output. One of these two layers is trained only examples from task 1 (documents) and the other is trained on only examples from task 2 (sentences). Both layers' outputs are subject to a sigmoid activation function that maps output values between 0 and 1 corresponding to predictions of non-protest or protest, respectively. Dropout of between 0.4 and 0.6 is applied between each layer (including the input layer) and values are chosen empirically using the \emph{dev} data set. The selected loss function is log loss and the model is fit with RMSProp~\cite{hinton:etal:slides}. The model architecture is shown in Figure~\ref{fig:task12}. 

The multitask nature of this model, having separate outputs for document and sentence-level predictions, allows the two tasks to jointly train the single LSTM layer; this effectively increases the training data size for this layer. Having two outputs allows the model to specialize for the two subtly-different tasks. Given a document, it may be the case that only a small portion of the document (a handful of words) refers to a protest. On the other hand, given a single sentence about a protest, it is likely that  a relatively larger portion of the words in that sentence refer to the protest in question. Task 1 requires that the model be sensitive to the small proportion of words indicative of a protest event in a larger document; task 2 is not necessarily so constrained. However, in hindsight, separating these tasks may not have been necessary: the document sub-model performs comparably to the sentence sub-model on sentence input and vice versa. 

\begin{figure}\centering
\includegraphics[width=1\linewidth]{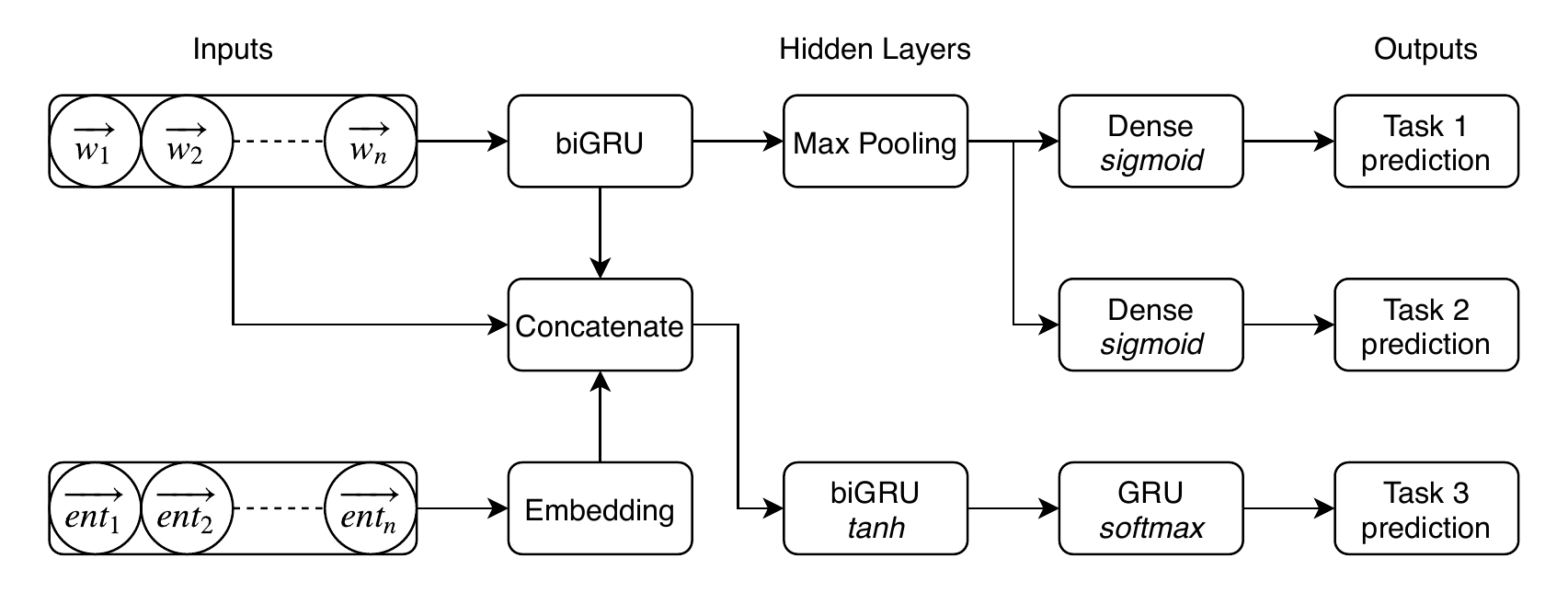}
\caption{Model architecture for task 3. Sequence input is shown on the left for word vectors $\vec{w_1}$ through $\vec{w_n}$ and corresponding one-hot encoded entity values $ent_1$ through $ent_n$.}\label{fig:task3}
\end{figure}

\subsection{Task 3 Model}

The model architecture for task 3 differs from that of tasks 1 and 2.\footnote{This is due, at least in part, to the fact that the competition was structured in such a way that tasks 1 and 2 were judged simultaneously and 3 was judged later. It is my believe that the model for task 3 would have fared similarly well had it diverged less from the model for tasks 1 and 2.} The input is still a time-ordered sequence of word vectors representative of a document or sentence. The LSTM layer has been replaced by a layer of 20 bidirectional gated recurrent units (GRU)~\cite{cho:etal:2014}. Instead of outputting the activation values of the GRU layer for the last sequence token, all activation values for the sequence are output. For the portions of the model corresponding to tasks 1 and 2, the output sequences from the GRU layer are flattened along the time axis by taking the maximum output value for each timestep. Aside from these minor changes and adjustments to dropout rates, the task 1 and task 2 sub-models are as described above. Task 3, the semantic role labeling task, requires a more complicated model. In particular, this sub-model consists of an additional bidirectional GRU layer with hyperbolic tangent activation and a subsequent unidirectional GRU layer that outputs a sequence of softmax-normalized predictions for each word's semantic role. The bidirectional GRU layer in this sub-model inputs not only the output sequence produced by the shared GRU layer but also inputs the original sequence of word embeddings as well as a sequence corresponding to the named entities identified in the input sequence. The three input sequences (shared GRU output, word vectors, and named entities) are concatenated word-for-word. Named entities are discovered using Spacy, a natural language processing module written in Python~\cite{spacy2}. Dropout of 0.25 is included between every layer of the task 3 model. The full model architecture is shown in Figure~\ref{fig:task3}.

\section{Results}

The two models are able to perform all three tasks at levels competitive with the reported results of similar research efforts.

The models were each trained for 100 epochs on consumer-grade hardware including a 6 core CPU and an NVIDIA 1070Ti GPU.\footnote{An epoch is defined as 20 batches per task where a batch is comprised of 128 training examples.} Training times were typically under 30 minutes. Models were written in Keras \cite{chollet:2015}, a machine learning library for Python that wraps TensorFlow~\cite{tensorflow:2015}. 

Due to the setup of the Lab, the true outcome values ($y$) for all \emph{test} and \emph{China} data sets are unavailable. Therefore, only the high-level summary metrics (F1 scores) provided as feedback by the ProtestNews Lab online evaluation system are reported for those data sets. More complete results (including precision and recall) are provided with respect to the \emph{train} and \emph{dev} data sets because they can be computed without the out-of-sample $y$ values.

While the models are able to generalize out-of-sample, their performance degrades noticeably as the task resolution becomes finer (from documents to sentences to words) and as the data transition from in-sample to validation, out-of-sample, and out-of-DGP. This is not surprising as the training data sets for tasks 1, 2, and 3 contain less overall information for each subsequent task (and, in the case of task 3, ask more of that limited amount of information). The decreasing performance from in-sample to out-of-sample data sets points to overfitting, a common problem in models with many parameters and one that can sometimes be remedied with additional training data and data augmentation techniques.

\begin{figure}
\centering
\subfigure[Task 1 (Document)]{\label{fig:task1confusion}\includegraphics[width=0.48\linewidth]{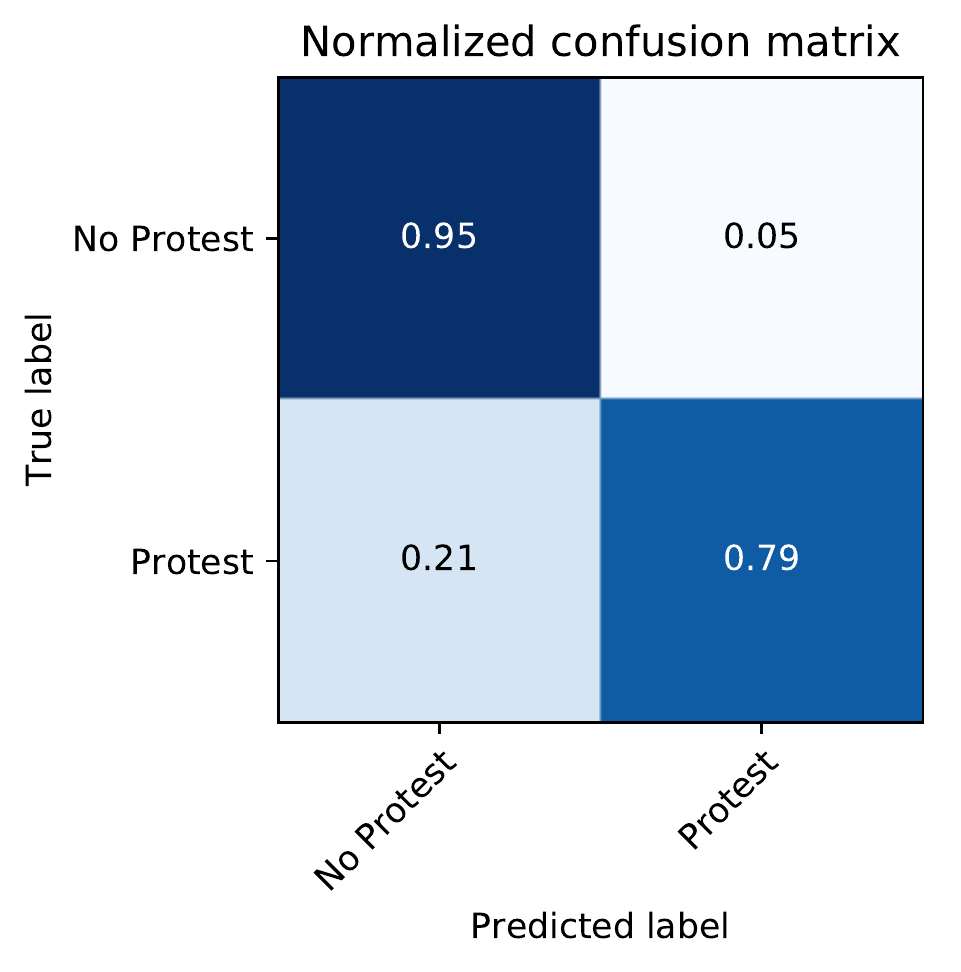}}
\hspace{0.01\linewidth}
\subfigure[Task 2 (Sentence)]{\label{fig:task2confusion}\includegraphics[width=0.48\linewidth]{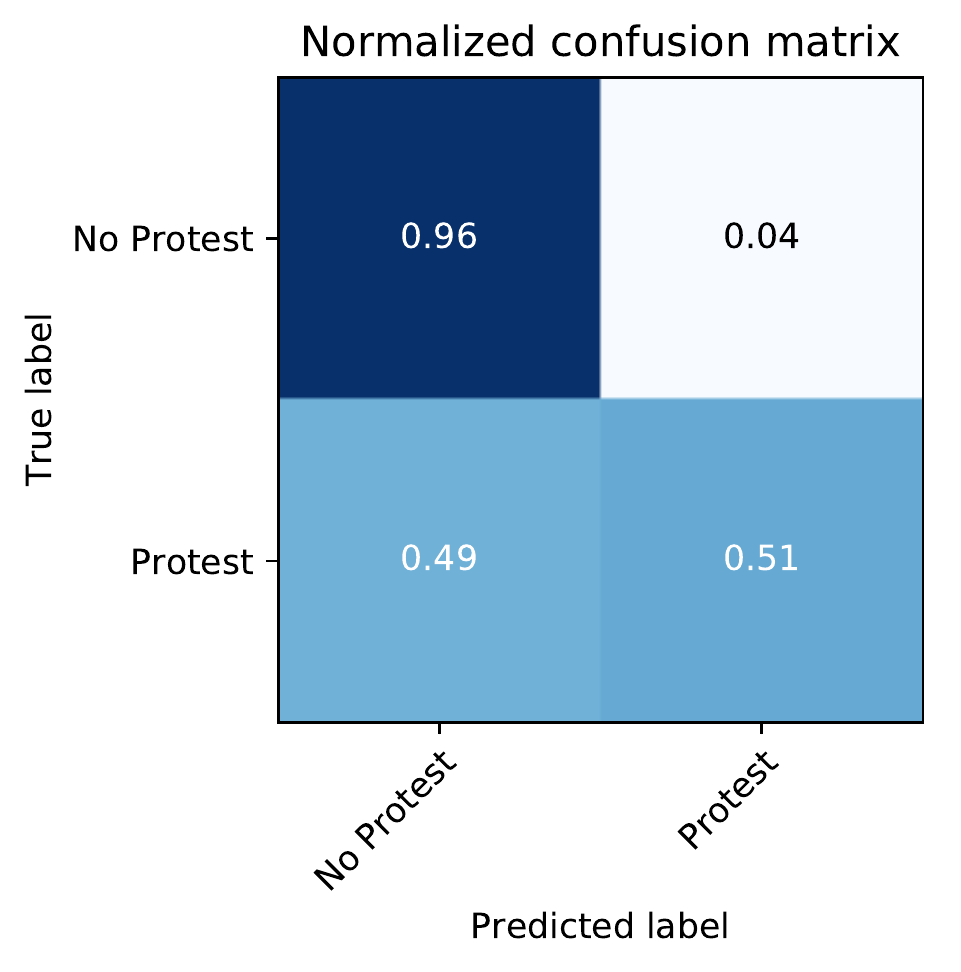}}
\caption{Model performance for tasks 1 and 2.}
\end{figure}

\subsection{Task 1 and 2 Results}

The multitask model for tasks 1 and 2 is able to classify documents with 92\% accuracy on the \emph{dev} set with an F1 score of 0.80.\footnote{Note that the \emph{dev} set was available at training time but was at no point provided to the model before inference was performed. Therefore it is out-of-sample but was available for hyperparameter tuning. Only limited results are available for the true out-of-sample datasets as these sets are held by the lab organizers.} As can be seen in Figure~\ref{fig:task2confusion}, the model correctly predicts 95\% of non-protest events and 79\% of protest events in the \emph{dev} set. In Table~\ref{tab:task12results} we can see the true out-of-sample F1 scores associated with the \emph{test} and \emph{China} sets are 0.84 and 0.66, respectively. This result on the \emph{test} set is encouraging because it is actually above the corresponding value on the \emph{dev} set, 0.80, and suggests that the model has not be overfit to the \emph{dev} set through hyperparameter selection. 


Model performance deteriorates somewhat for task 2, sentence classification. While the model accuracy is still high at 87\%, its precision has dropped markedly. In other words, the model is able to accurately classify non-protest event sentences (96\% accuracy) but only classifies protest events correctly 51\% of the time. This can be seen in Figure~\ref{fig:task2confusion}. In out-of-sample tests, the model achieves an F1 score of 0.66 on the \emph{test} set and 0.46 on the \emph{China} set.

\begin{table}\centering
\caption[Results for task 1 and 2 model]{Results for task 1 and 2 model.\footnotemark{}}\label{tab:task12results}
\begin{tabular}{@{\extracolsep{4pt}}lrrrrrr@{}}
\hline\hline
	& \multicolumn{3}{c}{Task 1} & \multicolumn{3}{c}{Task 2} \\ \cline{2-4}\cline{5-7}
	& Precision & Recall & F1 & Precision & Recall & F1  \\ \hline
Train & 0.93 & 0.84 & 0.88 & 0.63 & 0.82 & 0.71 \\
Dev  & 0.79 & 0.81 & 0.80 & 0.51 & 0.79 & 0.62 \\
Test  & -- & -- & 0.84 & -- & -- & 0.66  \\
China  & -- & -- & 0.66 & -- & -- & 0.46   \\ \hline\hline
\end{tabular}
\end{table}

\footnotetext{All values computed using contest-provided code. Precision and recall values were not provided by the contest organizers in the output of \emph{test} and \emph{China} data set evaluations and are therefore unavailable here. Unfortunately, the random seed values for the models trained and evaluated on the \emph{test} and \emph{China} data sets were lost and so the (\emph{train} and \emph{dev}) and (\emph{test} and \emph{China}) results are from two different model runs.}

\subsection{Task 3 Results}

The model that produced the Lab-submitted results for task 3 is actually capable of performing all three tasks, the results of which are shown in Table~\ref{tab:task3results}. However, I focus here on the results for semantic role labeling as this model was only evaluated on the \emph{test} and \emph{China} data sets for that particular task.  The precision, recall, and F1 scores shown here are multiclass weighted averages computed with a Lab-provided script. The unweighted average accuracy of this model on the \emph{dev} set is very high, 94\%, due largely to class imbalance. The model correctly predicts that most words in each sentence are not one of the selected roles. However, the model appears to generalize poorly: the F1 score for the \emph{train} data set is 0.82 but drops to 0.50, 0.52, and 0.39 for the \emph{dev}, \emph{test}, and \emph{China} data sets, respectively. This is indicative of a model that is overfit to the training data. 


\begin{table}\centering
\caption[Results for task 3 model]{Results for task 3 model.\footnotemark{}}\label{tab:task3results}
\begin{tabular}{@{\extracolsep{4pt}}lrrrrrrrrr@{}}
\hline\hline
	& \multicolumn{3}{c}{Task 1} & \multicolumn{3}{c}{Task 2} & \multicolumn{3}{c}{Task 3} \\ \cline{2-4}\cline{5-7}\cline{8-10}
	& Precision & Recall & F1 & Precision & Recall & F1 & Precision & Recall & F1 \\ \hline
Train & 0.99 & 0.98 & 0.98 & 0.87 & 0.97 & 0.91 & 0.83 & 0.81 & 0.82 \\
Dev  & 0.84 & 0.88 & 0.86 & 0.55 & 0.84 & 0.66 & 0.50 & 0.50 & 0.50 \\
Test  & -- & -- & -- & -- & -- & -- & 0.63 & 0.44 & 0.52 \\
China  & -- & -- & -- & -- & -- & -- & 0.54 & 0.31 & 0.39 \\ \hline\hline
\end{tabular}
\end{table}

\footnotetext{All values computed using contest-provided code. Due to time constraints imposed by the contest structure, performance was not evaluated for tasks 1 and 2 on the \emph{test} and \emph{China} data sets. Unfortunately, the random seed values for the models trained and evaluated on the \emph{test} and \emph{China} data sets were lost and so the (\emph{train} and \emph{dev}) and (\emph{test} and \emph{China}) results are from two different model runs.}

\begin{figure}
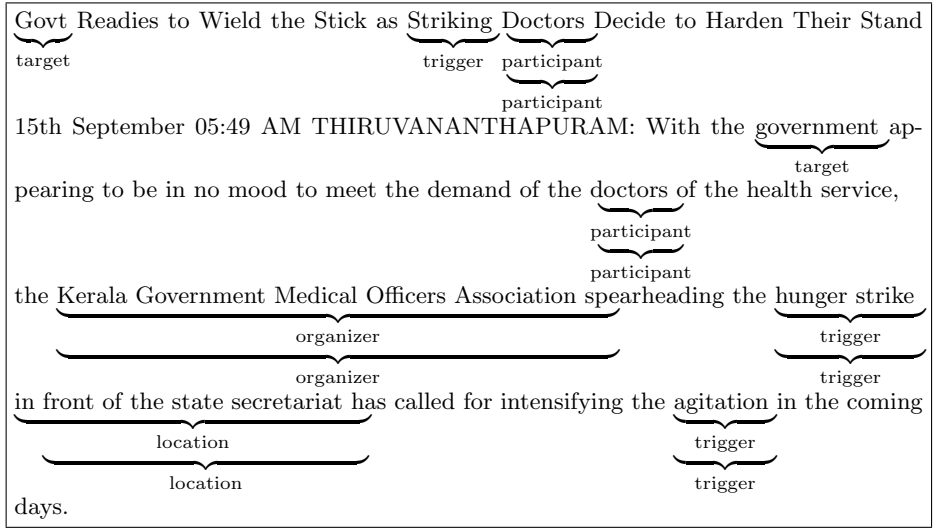

\framebox{\parbox{0.99\linewidth}{
\unote{11}Govt Readies to Wield the Stick as \unote{12}Striking \unote{1}\unote{13}Doctors Decide to Harden Their Stand 15th September 05:49 AM THIRUVANANTHAPURAM: With the \unote{14}government appearing to be in no mood to meet the demand of the \unote{3}\unote{15}doctors of the health service, \\the \unote{4}\unote{16}Kerala Government Medical Officers Association spearheading the \unote{5}\unote{17}hunger strike \\\unote{18}in \unote{6}front of the state secretariat has called for intensifying the \unote{7}\unote{19}agitation in the coming days.}}
\caption{An example of a task 3 \emph{dev} data set sentence with actual and predicted annotations . The top row of annotations are those labels that are provided with the data (i.e. ``true labels''). The bottom row of annotations are those that are predicted by the model. For example, ``in front of the state secretariat'' is coded as a \emph{location} in the \emph{dev} data but only the words ``front of the state secretariat'' are identified by the model as a \emph{location.}}\label{fig:annotations}
\end{figure}

An example of a task 3 \emph{dev} data set sentence with actual and predicted annotations is shown in Figure~\ref{fig:annotations}. This example illustrates five of the seven role categories and includes a target, participants, an organizer, triggers, and a location. For each row of text there are up to two rows of annotations. The top row of annotations represents the true role values provided by the Lab organizers. The bottom row of annotations are those predicted by the model.\footnote{In fact, the model must distinguish between the beginning token of a role phrase and the ``internal'' tokens. For example, ``Kerala Government Medical Officers Association'' would be annotated, word-for-word, ``B-organizer I-organizer I-organizer I-organizer I-organizer.'' These are omitted for clarity.}

\unotedef{1}{Doctors}{\scriptsize participant}{4ex}
\unotedef{3}{doctors}{\scriptsize participant}{4ex}
\unotedef{4}{Kerala Government Medical Officers Association}{\scriptsize organizer}{4ex}
\unotedef{5}{hunger strike}{\scriptsize trigger}{4ex}
\unotedef{6}{front of the state secretariat}{\scriptsize location}{4ex}
\unotedef{7}{agitation}{\scriptsize trigger}{4ex}
\unotedef{11}{Govt}{\scriptsize target}{0ex}
\unotedef{12}{Striking}{\scriptsize trigger}{0ex}
\unotedef{13}{Doctors}{\scriptsize participant}{0ex}
\unotedef{14}{government}{\scriptsize target}{0ex}
\unotedef{15}{doctors}{\scriptsize participant}{0ex}
\unotedef{16}{Kerala Government Medical Officers Association}{\scriptsize organizer}{0ex}
\unotedef{17}{hunger strike}{\scriptsize trigger}{0ex}
\unotedef{18}{in \unote{6}front of the state secretariat}{\scriptsize location}{0ex}
\unotedef{19}{agitation}{\scriptsize trigger}{0ex}

\begin{table}\centering
\caption{Role-wise performance on task 3 \emph{dev} set}\label{tab:roleperformance}
\begin{tabular}{@{\extracolsep{4pt}}lrrr@{}}
\hline\hline
 & Precision & Recall & F1 \\ \hline
 event time & 0.04 & 0.03 & 0.03 \\
 facility name & 0.13 & 0.07 & 0.09 \\
 location & 0.00 & 0.00 & 0.00 \\
 organizer & 0.52 & 0.43 & 0.47 \\
 participant & 0.65 & 0.60 & 0.63 \\
 place & 0.65 & 0.48 & 0.55 \\
 target & 0.13 & 0.41 & 0.20 \\
 trigger & 0.76 & 0.75 & 0.76 \\
 \hline
\end{tabular}
\end{table}

One-versus-all classification performance for the various role types is shown in Table~\ref{tab:roleperformance}. These metrics are evaluated on the out-of-sample \emph{dev} set. The model performs better on common role labels than less common labels; it achieves F1 scores greater than 0.5 on \emph{triggers}, \emph{participants}, and \emph{places}. The model fails to label any \emph{locations} correctly. This is probably due to the model's failure to recognize the prepositions preceding locations as the B tokens in the \emph{location} phrase. For example, ``in front of the state secretariat'' should be labeled ``B-loc, I-loc, I-loc, I-loc, I-loc, I-loc.'' Instead, the model predicts ``O, I-loc, I-loc, I-loc, I-loc, I-loc.'' Another example from the \emph{dev} set reads ``near a mosque'' and should be labeled ``B-loc, I-loc, I-loc.'' The model instead predicts ``B-fname, I-loc, I-loc,'' where ``fname'' represents the role ``facility name.'' 

The model for task 3 is also able to perform document and sentence classification. While \emph{test} and \emph{China} set results are unavailable for this model with respect to these two tasks, the model's performance on \emph{train} and \emph{dev} improves upon the results presented in Table~\ref{tab:task12results} across the board. Future work should determine whether this is due to the second model's ability to overfit to these tasks and data sets or due to the inclusion of task 3 data in the model's first GRU layer. One approach for exploring this is discussed in the paper's final section.

\subsection{Comparison to the State-of-the-Art}

H{\"u}rriyeto{\u{g}}lu et al. present preliminary findings for tasks 1 and 2~\cite{hurriyetoglu:etal:2019}. The results shown here compare favorably to the best of their models on both tasks.\footnote{The data set used in \cite{hurriyetoglu:etal:2019} is similar to, but may not be identical to, the data set used here. Therefore, caution should be taken when comparing the results between these two papers.} A model based on BERT~\cite{devlin:etal:2018}, for example, is reported to score F1=0.90 and F1=.64 on data sets roughly equivalent to \emph{test} and \emph{China} for task 1, just above and below the scores of 0.84 and 0.66 reported here. The authors report a high score of F1=0.56 on task 2 \emph{test} data achieved by a support vector machine model; this falls short of the bidirectional LSTM that scored F1=0.66. 

Previous studies have evaluated the performance of both human and machine-based coding for political event data. One of these reports that the ICEWS Jabari-NLP system achieves an average top-level event category precision of 75.6\% (document level). The authors further report that the system achieves average top-level event category recall values for documents and sentences of 65\% and 59\%, respectively~\cite{boschee:etal:2015}. This evaluation matches most closely with tasks 1 and 2 here and, in all cases, the above-presented protest models outperform the Jabari-NLP system results. Of course, the Jabari-NLP system was burdened with classifying 19 different event types while the task at hand represents only one. 

One previous study of undergraduate coders tasked with classifying top-level event categories found that the three coders achieved precision values of 39\%, 48\%, and 55\%~\cite{king:lowe:2003}. As was the case with the Jabari-NLP comparison, these particular precision values are not directly comparable to those reported for the protest models due to the fact that the human coders were provided a multiclass classification task, not binary. 

Using convolutional neural networks, Beieler \cite{beieler:2016} reports precision scores of 0.85 and 0.60 for \emph{QuadClass} classification on English and machine-translated texts, respectively, when word tokens are used. If character-based tokens are used, these scores increase to 0.94 (English) and 0.93 (native Arabic). However, the task presented is one of event classification conditional on the existence of an event in the text. This contrasts with the binary event/non-event objective of tasks 1 and 2. Nonetheless, these results point to a path forward for continued work on protest event detection via character-based models and convolutional neural networks.

\section{Discussion}

A drawback of the multitask RNN models used above is that they do not lend themselves to model interrogation -- they are typically viewed as black box models whose parameters resist simple interpretation. However, the addition of an attention layer to the input sequences would allow researchers to identify those input tokens (i.e. words) that contribute the most (or least) to a given prediction. Attention layers do this by masking some input tokens and not others, conditional on the input sequence itself. This would help to answer the question of which words contribute to accurate or inaccurate model predictions and whether those informative words differ from task to task.\footnote{While there does not appear to be a single best citation for attention neural networks, the earliest use of attention in RNN models may be \cite{mnih:etal:2014}.}

The models presented here make use of sub-word (i.e. character) information but only in the construction of word vectors from a pre-trained FastText model. By the time the sequences are input to the recurrent neural network models, the sub-word information has been aggregated to word-level tokens. Based on previous research that demonstrates the advantages of character-based models \cite{beieler:2016}, foregoing aggregation to the word vector level altogether may be beneficial. Instead, distributed character n-gram vectors could form the input sequences to a neural network classifier like those discussed above. This may, for example, allow the model to learn that n-gram vectors representing capitalized letters are more likely to occur in proper nouns, even if those proper nouns have never before been seen by the model. 

Finally, the CLEF 2019 ProtestNews Lab has provided the research community with a valuable ``ground truth'' data set on protest (and non-protest) events. The lack of hand-annotated and curated event data sets has made difficult the evaluation of event coding systems. Furthermore, due to copyright concerns that the ProtestNews organizers have cleverly overcome, previous event data sets have not published the underlying text data from which event records were derived.\footnote{The organizers provided a script that allowed participants to download the story data themselves from the original source websites.} Now that annotated text data are available, future solutions for deriving structured event records and their attributes from text should take advantage of this resources to evaluate their performance.

The results presented here indicate that supervised learning can achieve strong results in identifying politically-relevant events within unstructured text. However, the generalization of these models to out-of-sample data is imperfect; the ease with which neural network models like those used here can overfit to the training data means that care must be taken to ensure that the models continue to perform well on out-of-sample data. This is especially true if there is reason to believe that the out-of-sample data may represent a different data generating process than the in-sample data, as is the case here with the \emph{China} data set. Nonetheless, when sufficient training data are available (perhaps only a few thousand examples), supervised learning can play an important role in generating political event data. 

%
%
%
\bibliographystyle{splncs04}
\bibliography{master}

\end{document}